\documentclass{article}
\usepackage{spconf,amsmath,amssymb,graphicx,epsfig,mathptmx,times,multirow,mathrsfs,subcaption}
\usepackage{hyperref}
\usepackage{cite}
\usepackage{etoolbox}
\usepackage{color}
\title{DYNAMIC SUMMARY GENERATION FOR INTERPRETABLE MULTIMODAL DEPRESSION DETECTION}
\name{\begin{tabular}{c} 
    Shiyu Teng$^{1}$, Jiaqing Liu$^{1}$, Hao Sun$^{1,2}$, Yu Li$^{1}$, Shurong Chai$^{1}$, Ruibo Hou$^{1}$, Tomoko Tateyama$^{3}$ \\ 
    Lanfen Lin$^{2}$, Yen-Wei Chen$^{1,*}$\thanks{*Corresponding Author: Yen-Wei Chen (chen@is.ritsumei.ac.jp)}
\end{tabular}}

\address{
    $^1$College of Information Science and Engineering, Ritsumeikan University, Shiga, Japan\\
    $^2$College of Computer Science and Technology, Zhejiang University, Hangzhou, China\\
    $^3$Faculty of Engineering, University of the Ryukyus, Japan
}

\begin{document}
\maketitle

\begin{abstract}
Depression remains widely underdiagnosed and undertreated because stigma and subjective symptom ratings hinder reliable screening. To address this challenge, we propose a coarse-to-fine, multi-stage framework that leverages large language models (LLMs) for accurate and interpretable detection. The pipeline performs binary screening, five-class severity classification, and continuous regression. At each stage, an LLM produces progressively richer clinical summaries that guide a multimodal fusion module integrating text, audio, and video features, yielding predictions with transparent rationale. The system then consolidates all summaries into a concise, human-readable assessment report. Experiments on the E-DAIC and CMDC datasets show significant improvements over state-of-the-art baselines in both accuracy and interpretability.
\end{abstract}
\section{Introduction}
\label{sec:intro}

As a prevalent and severe mental disorder affecting hundreds of millions worldwide, the early, objective, and accurate assessment of depression is crucial for timely intervention \cite{cacheda2019early}. However, traditional diagnostic methods, which rely on clinical interviews and self-report questionnaires, have limitations such as strong subjectivity and high resource consumption, making them difficult to apply for large-scale screening \cite{newson2020heterogeneity}. Therefore, exploring automated and scalable computational methods to assist in depression assessment has become a pressing research direction.

\begin{figure}[!t]
\begin{center}
\includegraphics[width=0.75\linewidth]{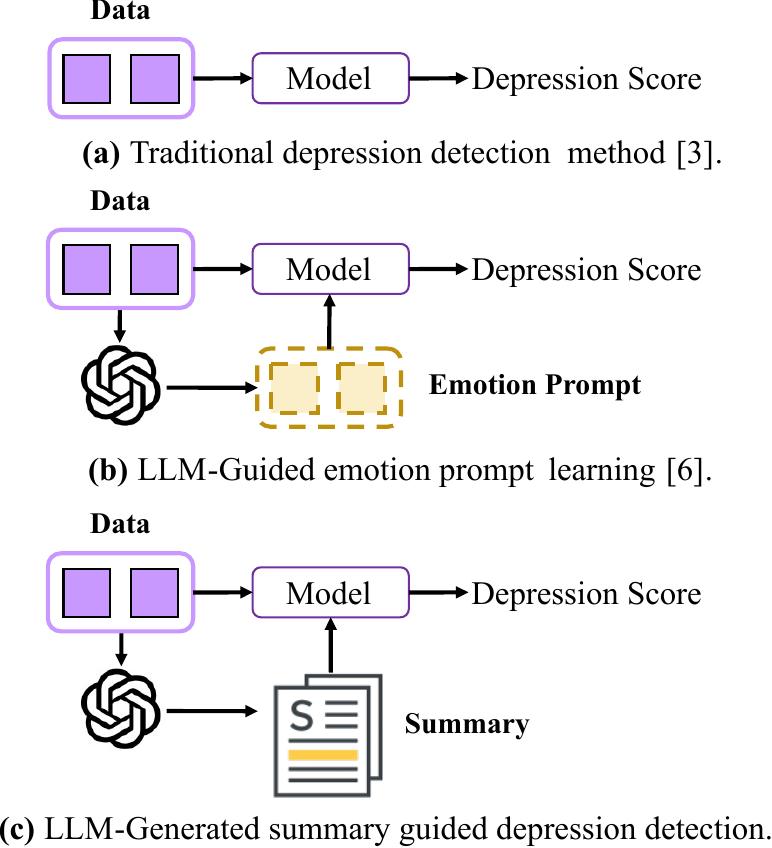}
\end{center}
\vspace{-10pt}
\caption{
An Overview of LLM-Driven Methods in Multimodal Depression Detection. 
} 
\label{fig:compare}
\end{figure}

\begin{figure*}[!t]
\begin{center}
\includegraphics[width=0.7\linewidth]{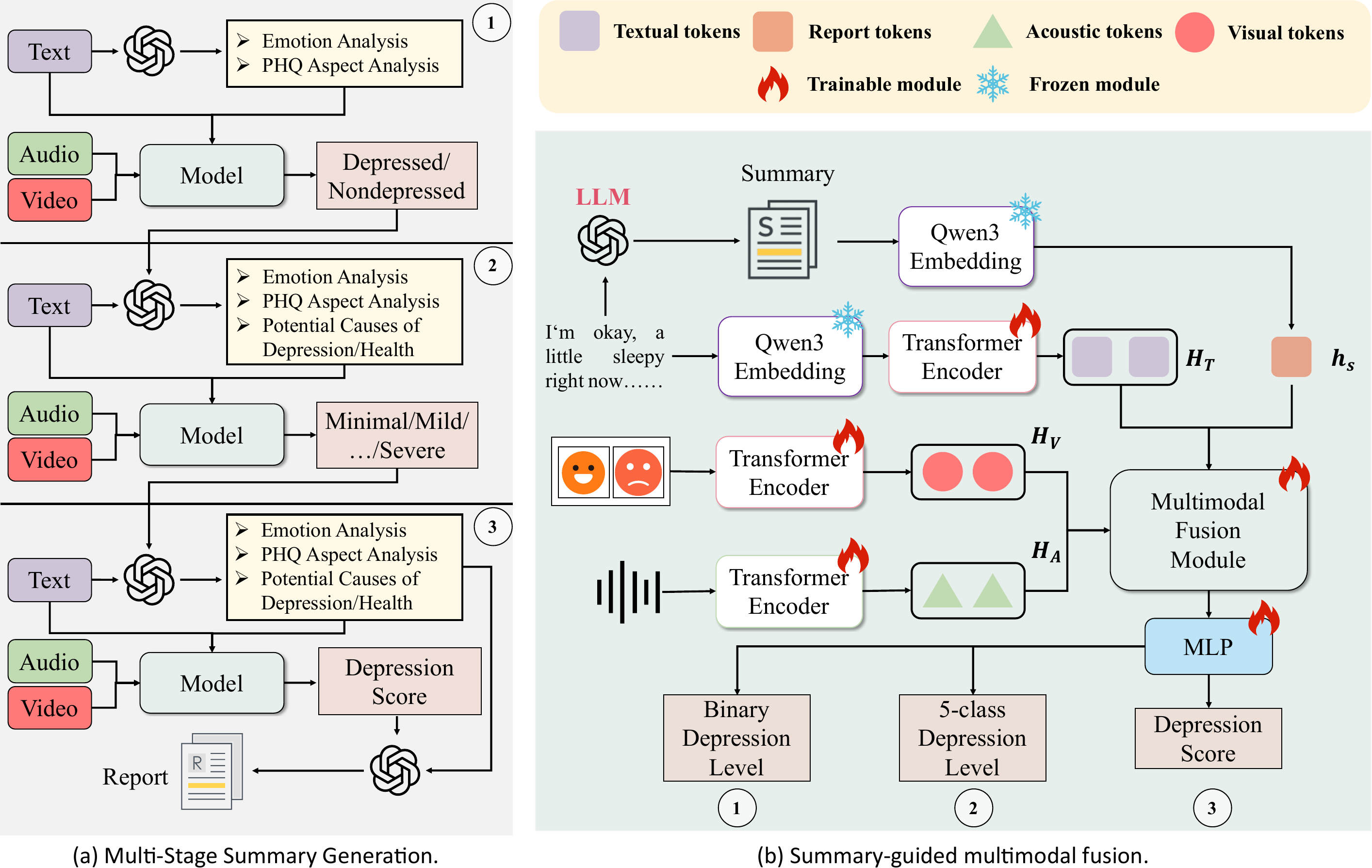}
\end{center}
\vspace{-10pt}
\caption{The overview of our proposed method. } 
\label{fig:overview}

\end{figure*}

Early computational pipelines learn directly from interview-derived multimodal signals (text, audio, video), as sketched in Fig. \ref{fig:compare}(a), and attempt to capture depression-related patterns from raw inputs \cite{sun2021multi, embcTENG}. Yet these signals contain substantial noise and non-diagnostic content; without explicit guidance, models often struggle to isolate clinically salient features, undermining robustness. More recently, LLM-based “emotion prompt” methods (Fig. \ref{fig:compare}(b)) leverage the link between affect and depressive symptoms to extract emotional attributes that steer detection \cite{ICASSP}. While beneficial, this line of work focuses narrowly on sentiment, overlooking additional clinical dimensions (e.g., potential causes), and thereby limits comprehensiveness.

To address the limitations of existing methods, we propose a novel LLM-Generated Summary Guided Depression Detection framework, with its core idea illustrated in Fig. \ref{fig:compare}(c). Our key innovation lies in a multi-stage, coarse-to-fine summary generation process. Instead of creating a single, static analysis, our framework instructs the LLM to generate progressively more detailed summaries tailored to each stage of our hierarchical detection pipeline—from an initial analysis for binary classification to a more nuanced summary incorporating potential causes for fine-grained score regression. This dynamic, staged summary provides far more comprehensive and context-aware guidance than a static "emotion prompt." By injecting this evolving summary as a guiding signal, our model can more intelligently locate, correlate, and integrate subtle cues across all modalities, thereby achieving a deeper understanding and more accurate assessment. Evaluations on the Extended Distress Analysis Interview Corpus (E-DAIC) \cite{gratch-etal-2014-distress} and Chinese Multimodal Depression Corpus (CMDC) \cite{CMDC} demonstrate that these improvements lead to more robust and accurate depression detection, outperforming current state-of-the-art methods.

\section{Method}
\label{sec:method}
Figure~\ref{fig:overview} overviews our summary-augmented multimodal framework. The pipeline comprises three stages: binary screening (Stage 1), five-class severity classification (Stage 2), and continuous regression (Stage 3). At each stage, an LLM generates a concise text summary, which is embedded alongside raw text, audio, and video by dedicated encoders. A multimodal fusion module integrates these embeddings, and an MLP predictor outputs the stage-specific depression estimate. After Stage 3, the final continuous score is combined with all stage-wise LLM summaries to produce a clinically oriented report.

\vspace{-5pt}
\subsection{Multi\mbox{-}Stage Summary Generation}
\label{sec:summary}

Figure~\ref{fig:overview}(a) illustrates the three progressively granular stages. In Stage 1 (binary screening), a pretrained LLM (GPT-o3 \cite{achiam2023gpt}) is prompted to produce a brief summary focused on emotional state and symptoms related to the Patient Health Questionnaire-8 (PHQ-8)\cite{PHQ}, a standard tool assessing depression across eight domains. We adopt o3 because it exhibits strong psychology-understanding performance on HealthBench \cite{arora2025healthbench}, indicating clinically oriented comprehension and suitability for our setting. This summary, together with audio and video features, is used to predict whether the individual is depressed or not. In Stage 2 (five-class classification), the Stage 1 label is included in the prompt, and the LLM returns a more detailed summary that incorporates inferred potential causes of depression as well as indicators of mental well-being, facilitating five-level severity assignment. In Stage~3 (regression), the prompt requests a comprehensive clinical note that integrates emotional cues, inferred causes, and a detailed PHQ-8 aspect analysis; the LLM explicitly evaluates the user against each of the eight PHQ-8 domains, enabling prediction of a continuous depression score. This score is then combined with the stage-wise summaries to form a structured, human-readable report that provides an interpretable rationale for all predictions.


\vspace{-5pt}
\subsection{Summary-guided multimodal fusion  \label{subsec:embedding}}

We use the Transformer\cite{transformer} to extract audio and video features, while a pre-trained Qwen3 Embedding\cite{zhang2025qwen3embeddingadvancingtext} model is employed to extract text and summary features. The extracted features can be mathematically represented as follows:
\begin{equation}
\left\{
\begin{aligned}
H_{T} &= \text{Transformer}\left(\text{Qwen3Embedding}\left(T \right)\right), \\ h_{S} &= \text{Qwen3Embedding}\left(S \right), \\
H_{A} &= \text{Transformer}\left(A \right), \quad H_{V} = \text{Transformer}\left(V \right),
\end{aligned}
\right.
\end{equation}
where $T\in \mathbb{R}^{L_t\times d_t}$, $S\in \mathbb{R}^{L_{s}\times d_t}$, $A\in \mathbb{R}^{L_a\times d_a}$, $V\in \mathbb{R}^{L_v\times d_v}$ denote the input textual, summary, audio, and video data respectively, where $d_{m}$ is the embedding dimension, and $L_{m}$ is the number of tokens in the input, m is modality. To enable joint fusion, we use a modality-specific MLP to project all embeddings into a unified dimension $d$.

Given the modality‐specific embeddings $H_{T}\!\in\!\mathbb{R}^{L_{t}\times d}$, $H_{A}\!\in\!\mathbb{R}^{L_{a}\times d}$, $H_{V}\!\in\!\mathbb{R}^{L_{v}\times d}$, and the summary vector $h_{S}\!\in\!\mathbb{R}^{d}$, our fusion module proceeds in three steps: report–guided gating, bidirectional cross–attention, and global aggregation.

\paragraph*{1) Report–guided gating}
A lightweight MLP takes the summary token and produces
modality–wise gate scalars
$g_{T},g_{A},g_{V}\!\in\!(0,1)$:
\begin{equation}
[g_{T},g_{A},g_{V}]^\top=\sigma\!\left(W_{g}h_{S}+b_{g}\right),
\end{equation}
where $W_{g}\!\in\!\mathbb{R}^{3\times d}$ and $b_{g}\!\in\!\mathbb{R}^{3}$ are trainable
and $\sigma(\cdot)$ is the sigmoid function.
The gated features are
$\tilde{H}_{T}=g_{T}H_{T}$,
$\tilde{H}_{A}=g_{A}H_{A}$,
$\tilde{H}_{V}=g_{V}H_{V}$.
We concatenate them to form
$H_{\mathrm{cat}}\!=\![\tilde{H}_{T};\tilde{H}_{V};\tilde{H}_{A}]
\in\mathbb{R}^{L\times d}$,
where $L=L_{t}+L_{v}+L_{a}$.

\paragraph*{2) Bidirectional cross–attention and attention pooling}
Two multi–head attention (MHA) blocks create reciprocal context:
\begin{align}
z_{1}&=\mathrm{MHA}(q=h_{S},\;k=H_{\mathrm{cat}},\;v=H_{\mathrm{cat}}),\\
z_{2}&=\mathrm{MHA}(q=H_{\mathrm{cat}},\;k=h_{S},\;v=h_{S}),
\end{align}
with $z_{1},z_{2}\!\in\!\mathbb{R}^{d}$.  In parallel, we prepend a
\texttt{[CLS]} token to $H_{\mathrm{cat}}$ and pass it through a
transformer encoder, obtaining a global representation
$H_{\mathrm{CLS}}\!\in\!\mathbb{R}^{d}$.

\paragraph*{3) Global aggregation}
The final multimodal embedding is the concatenation
\begin{equation}
H_{\mathrm{fusion}}=[h_{S}; H_{\mathrm{CLS}};\,z_{1};\,z_{2}]
\in\mathbb{R}^{4d},
\end{equation}
which is fed to an MLP predictor to output the stage-specific
depression target (binary label, five-class label, or continuous score).

\subsection{Loss function  \label{subsec:loss}}

We adopt a stage-wise objective aligned with the three-stage pipeline: the two classification stages—binary screening (Stage 1) and five-class severity classification (Stage 2)—are optimized with cross-entropy (CE), while the continuous regression stage (Stage 3) is optimized with the Concordance Correlation Coefficient (CCC) loss \cite{lawrence1989concordance}:
\begin{equation}
\mathcal{L}_{\mathrm{ccc}} = 1 - \frac{2 \, r \, s_{t} \, s_{p}}{s_{t}^2 + s_{p}^2 + (m_{t} - m_{p})^2},
\vspace{-5pt}
\end{equation}
where \(r\) denotes the Pearson correlation coefficient, \(s_{t}\) and \(s_{p}\) represent the standard deviations, and \(m_{t}\) and \(m_{p}\) denote the means of the target (\(t\)) and predicted (\(p\)) values, respectively.

\section{EXPERIMENTS\label{EXPERIMENTS}}

\subsection{Datasets}

\textbf{E-DAIC}\cite{gratch-etal-2014-distress} is a standard multimodal benchmark dataset for computational psychiatry, released as part of AVEC 2019\cite{ringeval2019avec}. This dataset consists of video, audio, and textual modalities, with accompanying annotations such as PHQ-8 scores~\cite{PHQ}, interview identifiers, depression classifications, and participant gender. E-DAIC includes 275 interview samples, which are split into training (163), development (56), and test (56) sets.

\textbf{CMDC} \cite{CMDC} is a benchmark dataset consisting of 78 clinical interviews, each containing synchronized audio, text transcripts, and visual features. Participants include both patients with major depressive disorder and healthy controls. Interviews follow a fixed twelve-question format aligned with the PHQ-9 scale.

\subsection{Implementation details} 
Batch size is set to 24. Learning rates are $7\times10^{-4}$ for E-DAIC and $5\times10^{-4}$ for CMDC, decayed by a factor of 0.1 every 50 epochs. Training runs for 150 epochs. All experiments are conducted on two RTX5090 GPUs using PyTorch with the AdamW optimizer\cite{loshchilov2019decoupled}.

\subsection{Evaluation metrics}

In our assessment of the model's performance for depression detection, we focus on different metrics for each dataset. For E-DAIC, we use the CCC~\cite{lawrence1989concordance} and Mean Absolute Error (MAE). For CMDC, we use Root Mean Squared Error (RMSE) and MAE.


\begin{table}[!htbp]
\vspace{-5pt}
\centering
\caption{The results on the test set of E-DAIC\cite{ringeval2019avec} dataset. $\downarrow$: the lower the better; $\uparrow$: the higher the better.}
\begin{tabular}{ccc}
\hline
Method  & CCC $\uparrow$   & MAE $\downarrow$  \\
\hline
AVEC2019 Baseline \cite{ringeval2019avec}   & 0.111 & 6.37   \\
AFT \cite{sun2021multi} & 0.443 & 5.66 \\
Wu \textit{et al.} \cite{Climate_and_Weather}  & 0.451 & 5.44 \\
Teng \textit{et al.} \cite{teng2024icce}  & 0.466 & 5.21 \\
TensorFormer \cite{tensorformer}  & 0.493 & 5.13  \\
 MIMRL \cite{sunMutual} & 0.580 & 4.36 \\ 
CubeMLP \cite{Cubemlp}  & 0.583 & 4.37 \\ 
Van Steijn \textit{et al.} \cite{Van}  & 0.62 & - \\ 
Yuan \textit{et al.} \cite{yuan2022depression}  & 0.676 & 3.98 \\
EP \cite{ICASSP} & 0.688 & 3.89 \\
GPT-o3 zero shot \cite{achiam2023gpt} & 0.698 & 3.67 \\
Ours & \textbf{0.717} & \textbf{3.32} \\
\hline
\end{tabular}
\label{table:comparison}
\vspace{-5pt}
\end{table}

\begin{table}[!htbp]
\centering
\caption{The results on the test set of CMDC\cite{CMDC} dataset.}
\begin{tabular}{ccccc}
\hline
Method  & RMSE $\downarrow$   & MAE $\downarrow$  \\
\hline
Bi-LSTM \cite{CMDC}  & 5.67 & 4.55 \\
MulT \cite{CMDC}  & 5.61 & 4.32 \\
PIE \cite{PIE}   & 5.56 & 4.58   \\
Ours  & \textbf{3.81} & \textbf{2.79} \\
\hline
\end{tabular}
\label{table:comparison_eatd}
\vspace{-10pt}
\end{table}

\subsection{Comparison with State-of-the-art methods}


Tables~\ref{table:comparison} and~\ref{table:comparison_eatd} compare our approach with recent methods on E-DAIC~\cite{gratch-etal-2014-distress} and CMDC~\cite{CMDC}. Relative to EP~\cite{ICASSP}, which injects emotion prompts to guide prediction, our framework integrates multiple clinically relevant components—emotion, PHQ aspects, and potential causes—within a unified multimodal pipeline. Although our generation module leverages GPT-o3~\cite{achiam2023gpt}, the full model surpasses the GPT-o3 zero-shot baseline, yielding higher CCC and lower MAE on E-DAIC. On CMDC, it attains an RMSE of 3.81 and an MAE of 2.79. Jointly embedding emotion, PHQ-aspect, and causal information enables the model to capture a broader spectrum of depression-related cues, improving precision, robustness, and interpretability across both datasets.

\vspace{-3pt}
\subsection{Ablation study}

We conduct three groups of ablations on the E-DAIC test set~\cite{ringeval2019avec} to quantify the contribution of modalities, fusion modules, and summary composition.

Table~\ref{table:Modality} isolates modality effects for text (T), audio (A), video (V), and the LLM-generated summary (S). Text alone is strongest among single modalities; audio and video contribute less in isolation. Adding the summary yields a marked gain, and fusing all four sources achieves the best performance, underscoring the benefit of holistic integration.

Table~\ref{table:Fusion} evaluates fusion components. Report-guided gating (Gate), bidirectional cross-attention (BCA), and attention pooling (AP) provide incremental improvements, with all modules combined delivering the highest accuracy. This highlights the roles of adaptive weighting, cross-modal alignment, and global aggregation.

Table~\ref{table:prompt_LLM} examines the semantic richness of the LLM-generated summary across stages. Incorporating progressively richer clinical semantics from stage 1 to stage 3 consistently improves performance, aligning with the coarse-to-fine design.





\begin{table}[!htbp]
\centering
\caption{Ablation on modality contributions using E-DAIC~\cite{gratch-etal-2014-distress}. “T”: text, “A”: audio, “V”: video, “S”: LLM-generated summary.
}
\begin{tabular}{ccccccc}
\hline
T & A & V & S &CCC $\uparrow$ & MAE $\downarrow$   \\
\hline
 $\surd$&  &  & &0.577 & 4.05 \\
 &$\surd$  &  & & 0.288 & 6.41 \\
 &  & $\surd$ & &0.323 & 5.76 \\
 &  &  &  $\surd$&0.653 & 3.63 \\
$\surd$ & $\surd$ &  & & 0.595 & 3.97 \\
 $\surd$& $\surd$ & $\surd$ & & 0.602 & 3.95 \\
$\surd$ & $\surd$ & $\surd$ &$\surd$& \textbf{0.717} & \textbf{3.32} \\
\hline
\end{tabular}
\label{table:Modality}
\end{table}

\begin{table}[!htbp]
\centering
\caption{
Ablation on fusion modules for E-DAIC~\cite{gratch-etal-2014-distress}. “Gate”: report-guided gating; “BCA”: bidirectional cross-attention; “AP”: attention pooling.
}
\begin{tabular}{cccccc}
\hline
Gate & BCA & AP & CCC $\uparrow$ & MAE $\downarrow$   \\
\hline
 &  &  & 0.677 & 3.88 \\
$\surd$ &  &  & 0.688 & 3.84 \\
$\surd$ & $\surd$ &  & 0.703 & 3.51 \\
$\surd$ & $\surd$ & $\surd$ & \textbf{0.717} & \textbf{3.32} \\
\hline
\end{tabular}

\label{table:Fusion}

\end{table}

\begin{table}[!htbp]
\centering
\caption{
Effect of summary composition across stages on E-DAIC~\cite{gratch-etal-2014-distress}. 
}
\begin{tabular}{lcc}
\hline
Summary Composition                        & CCC $\uparrow$   & MAE $\downarrow$  \\
\hline
Stage 1                   & 0.683   & 3.61     \\
Stage 2        & 0.708   & 3.55     \\
Stage 3   & \textbf{0.717}   & \textbf{3.32}     \\
\hline
\end{tabular}
\label{table:prompt_LLM}
\vspace{-10pt}
\end{table}

\vspace{-5pt}
\section{CONCLUSION}
\label{sec:conclusion}
In this work, we proposed a novel summary-guided multimodal depression detection framework to address the limitations of methods that rely on fragmented or narrowly-focused guidance. Our core innovation is a multi-stage, coarse-to-fine summary generation process that provides dynamic and holistic semantic guidance by integrating emotional cues, PHQ-related symptoms, and potential causal factors. Evaluations on E-DAIC and CMDC datasets demonstrate that our approach significantly outperforms state-of-the-art methods, showcasing its robustness and effectiveness in advancing comprehensive, clinically-aligned depression detection.

\section{ACKNOWLEDGMENT}

This work was supported in part by the Grant-in-Aid for Scientific Research from the Japanese Ministry of Education, Culture, Sports, Science and Technology (MEXT) under Grant No. 20KK0234; by JSPS KAKENHI Grant No. JP23K16909; by JST CREST (JPMJCR25T4) and JST BOOST (JPMJBS2428); and by the Natural Science Foundation of Zhejiang Province (Grant No. LZ22F020012).

\section{COMPLIANCE WITH ETHICAL STANDRDS}

This study is retrospective and uses human-subject data that are publicly available from the DAIC-WOZ database released by USC ICT and the Chinese Multimodal Depression Corpus released on IEEE DataPort. Ethical approval was not required as confirmed by the license attached with the open access data.


\bibliographystyle{IEEEbib}
\bibliography{new}

\end{document}